# A Census-Based Genetic Algorithm for Target Set Selection Problem in Social Networks

**Md. Samiur Rahman[1], Mohammad Shamim Ahsan[2*], Cheng-Wu Chen[3†], Vijayakumar Varadarajan[4, 5, 6†]**

[1]Department of Computer Science and Engineering, Presidency University, Bangladesh
[2*]Department of Computer Science and Engineering, United International University, Bangladesh
[3]National Kaohsiung University of Science and Technology, Taiwan
[4]Program Leader, Engineering, Ajeenkya D Y Patil University, India
[5]University of Technology Sydney, Australia
[6]Swiss School of Business and Management, Switzerland
*Corresponding author(s). E-mail(s): ahsan@cse.uiu.ac.bd;
Contributing authors: samiurr@pu.edu.bd; chengwuchen@nkust.edu.tw; dean.international@adypu.edu.in
†These authors contributed equally to this work.

**Abstract**

This paper considers the Target Set Selection (TSS) Problem in social networks, a fundamental problem in viral marketing. In the TSS problem, a graph and a threshold value for each vertex of the graph are given. A minimum size vertex subset needs to be found to "activate" such that all graph vertices are activated at the end of the propagation process. Specifically, we propose a novel approach called "a census-based genetic algorithm" for the TSS problem. In our algorithm, the concept of a census is utilized to gather and store information about individuals in a population and collect census data from the individuals constructed during the algorithm's execution so that greater diversity and avoiding premature convergence can be achieved at locally optimal solutions. Specifically, two distinct census information have been used: (a) for individuals, the algorithm stores how many times it has been identified during the execution; (b) for each network node, the algorithm counts how many times it has been included in a solution. The proposed algorithm can also be self-adjusted by using a parameter specifying the aggressiveness employed in each reproduction method. Additionally, the algorithm is designed to run in a parallelized environment to minimize the computational cost and check individual's feasibility. Moreover, our algorithm finds the optimal solution in all cases while experimenting on random graphs. Furthermore, the proposed algorithm has been employed on 14 large graphs of real-life social network instances from the literature, improving around 9.57 solution size (on average) and 134 vertices (in total) compared to the best solutions obtained in previous studies.

## 1 Introduction

Nowadays, social media has become the central platform to connect, communicate, access, and share information smoothly and continuously with friends, followers, and business partners, largely influencing our everyday lives. Most importantly, people excessively tend to follow and lean toward their followers' and followees' thoughts and activities. By observing these trends, significant interests have arisen from algorithmic researchers to understand the dynamics of adoption within a social network, which may result in compelling clues to better marketing strategies [21]. Consequently, a diverse range of practical problems in social media, such as target set selection, volume prediction [46], source prediction [47], buzz prediction [45], link detection [48], etc., have been studied to understand better how this virtual platform behaves and propagates information. In this work, we focus on exploring the well-known target set selection problem.

### 1.1 Target Set Selection (TSS) problem

Given an undirected graph, $G = (V, E)$ and a vector $R$ of nonnegative integers such that $0 \leq R[v] \leq dev(v)$ for every $v \epsilon V$, the Target Set Selection Problem (TSS) consists of finding a minimum set $D \subseteq V$ that satisfies the following domination rule: every vertex $v \epsilon V$ must be included in the set $D_{n-1}$, where $D_0 = D$ and $D_{i+1} = D_i \cup \{ v \epsilon V \setminus D_i : | N(v) \cap D_i | \geq R[v] \}$. Trivially, every $v \epsilon V$ satisfies the domination rule. Also, repeated applications of the domination rule determine an irreversible process that stabilizes at some iteration, i.e., $D_{i+1} = D_i$ for some $i < n$. The values $R[v]$ are called the vertex requirements. A solution $D$ of the TSS is said to iteratively dominate (or simply dominate) the input graph $G$.

Some interesting variations of the TSS are defined by setting an upper bound on the number of iterations. The Vector Domination Problem [8] aims at finding a minimum $D$ such that every vertex must be dominated in a single iteration, i.e., for all $c$, either $v \epsilon V$ or $| N(v) \cap D_i | \geq R[v]$. In this context, the case $R[v] = k$ for every $v \epsilon V$ is precisely the well-known $k$-Dominating Set Problem, which is NP-hard for general graphs even when $k = 1$; the case $R[v]two = [\deg[v]/2]$ models' monopolies, consensus, and voting problems [9]. Another interesting variation is the Vaccination Problem [10], which aims to determine a minimum set that prevents any other set from dominating the entire graph. Those sets are called vaccines since they can stop the propagation of diseases or influence.

### 1.2 Related Works

The Target Set Selection (TSS) problem is a fundamental problem in social network analytics. Kempe et al. [22, 23] were the first to study problems of influence in networks from an algorithmic point of view. Specifically, their interests lie in networks with randomly chosen thresholds. TSS has been widely studied in literature, and some variants were introduced in [3, 19, 20, 22 - 27]. In the field of approximate solutions, the prominent works are Chen [4, 19], Dinh et al. [12], and Cordasco et al. [6, 7]. It is often used to computationally model the spread of diseases, influence, and opinions [1, 2]. Many works addressed this problem in instances that map social networks [1, 3-7]. One interesting problem in this universe is the propagation of fake news on social networks. It determines how many initial users (influencers) must take an opinion to ensure that it will propagate to a large network subgraph. Dreyer and Roberts [3] proved that the TSS remains NP-hard when all the vertex requirements are equal to a constant $k \geq 3$. In a later work, Centeno et al. [11] extended this NP-hardness result to $k=2$. In the same work, Centeno et al. [11] also described a polynomial-time algorithm for some families of graphs, like block graphs. The case $k=1$ is trivial: if $G$ is connected, any singleton is a solution of the TSS. Chen [19] showed that minimizing the target set size is APX-hard.

Raghavan and Zhang [39] proposed a linear-time dynamic programming algorithm for the weighted target set selection (WTSS) problem on trees and cycles. Later, they developed a branch-and-cut approach to solve the WTSS problem on arbitrary graphs [43]. Günneç et al. [44] studied social networks' least cost target set selection problem. They proposed a greedy algorithm and dynamic programming algorithm to solve the problem for the tree structure network. In the context of majority-based distributed system networks, TSS has been explored in [30 - 38], where white and black nodes were considered excellent and faulty, respectively. Chiang et al. [29] determined optimal target sets for various honeycomb networks such as honeycomb mesh, torus, rectangular torus, rhombic torus, generalized honeycomb rectangular torus, planar hexagonal grid, cylindrical hexagonal grid, and toroidal hexagonal grid under a strict majority threshold. Shakarian et al. [40] presented a heuristic and tested it on various real-world graphs. Later, Cordasco et al. [41] showed that a heuristic could provide an optimal solution for a particular case of the WTSS problem on complete graphs. In a parameterized setting, Hartmann [42] gave an FPT algorithm for TSS for the combined clique-width and maximum threshold, where the algorithm's time complexity grew only single-exponentially in the parameters.

### 1.3 Objectives and Contributions

This work proposes a novel genetic algorithm approach to address the target set selection problem. A genetic algorithm is a biology-inspired method that simulates several generations of a population formed by distinct individuals, where an individual is associated with a solution to the problem. The reproduction process creates new generations of individuals formed by permutations of the genes of the previous generation. At the end of this process, the best individual corresponds to the best solution found. Since the TSS is NP-hard for general graphs, intuitively, an alternative is to obtain good solutions in a reasonable time using metaheuristics, especially genetic algorithms. Our approach is called "a census-based genetic algorithm" that uses census-data information. Censuses are carried out to acquire and store information about the population at a given time. In our algorithm, census data are gathered and used to guide it to finer ways of evaluating solution quality and ensuring greater solution diversity. To do this, two types of census data are stored: S-Census and V-Census. The census-based approach maintains diversity in the genetic algorithm through a global activation frequency mechanism. The census operator computes the frequency of how often each node appears in the target sets across the entire population. Thus, it outlines a global perspective on which nodes are frequently or infrequently selected. Nodes in the network with lower activation frequencies are given higher probabilities to be included during mutation and crossover. This ensures the exploration of underrepresented nodes and prevents premature population convergence on a limited subset of nodes. The census operator ensures that the algorithm explores various areas of the solution space, concentrating on promising regions by directing the search toward frequently activated nodes (exploitation) and non-frequently activated nodes (exploration).

In addition to using census data information, a self-adaptive capability is also implemented, allowing the algorithm to adjust the aggressiveness of each operation to escape from local optimal solutions. Determining if a set D is a feasible solution to the TSS requires a high computational cost, especially in significant instances, and this is a critical point since a genetic algorithm creates many individuals over the generations. Our algorithm is designed to run in a parallelized environment to minimize the impacts of a feasibility check. The parallelization strategy consists of performing each feasibility check in a different processor, significantly reducing the time needed to complete each generation. To better exploit the environment, the reproduction methods and part of the fitness evaluation are performed in parallel, where processors run independently, except when they need to consolidate a new generation cooperatively. Consequently, our algorithm performs better than the existing well-known solutions by Cordasco et al. [6, 7] for random graphs and appropriate for any network graphs, ranging from simple to dynamic ones. In summary, the following contributions have been made:

- A census-based genetic algorithm integrated with self-adaptive capability is proposed to avoid being stuck at local optima. Specifically, the algorithm utilizes census data as a subordinate heuristic to guide for improving quality as well as diversity of the solutions.
- Additionally, each feasibility check is conducted separately and parallelly in different processors. As a result, the computational overhead has been reduced substantially which overcomes limitations of the classic genetic algorithm.

- Moreover, in experiments on 14 large real-life social networks, the proposed algorithm improves previously best-known solutions in almost all cases, with an average improvement of 9.57 vertices and a total improvement of 134 vertices.

The remainder of this work is organized as follows. In Section 2, some related terminologies are introduced. In Section 3, the proposed genetic algorithm is explained in detail. In Section 4, computational experiments are presented; first, a description is provided on how the values of each algorithm parameter influence its execution, and next, the results are shown on random instances and social network instances available in the literature [13, 14]. Finally, in Section 5, we conclude.

## 2 Preliminaries

In this section, a few concepts of meta-heuristics algorithms are described, and some terminologies related to genetic algorithms.

### 2.1 Meta-heuristics Algorithms

Metaheuristics is a general exploration (or diversification) method that applies to many different problems in the same way by visiting regions of the search space that is not already seen. In general terms, metaheuristics are approximation algorithms that provide good or acceptable solutions within an acceptable computing time that cannot be obtained with more specialized techniques, such as brute-force, linear programming, randomization, quantum computation, exact algorithms, etc., but do not give formal guarantees about the quality of the solutions [49]. Among the metaheuristics, simulated annealing (Kirkpatrick et al. [50]), ant colony method (Dorigo, Maniezzo, and Colorni [51]), particle swarm optimization (Eberhart and Kennedy [52, 53]), and evolutionary algorithms: genetic algorithms (J. Holland [54]), evolutionary programming (L. Fogel [55]) are well-known and established. *Simulated Annealing* is a method inspired by the physical metallurgy process and uses corresponding terminologies. First, an arbitrary admissible initial configuration and initial temperature (usually high temperature) are chosen. At a high temperature, the system is allowed to explore randomly many possible configurations, hence, accessible states. Then, the *annealing* process begins, that is, to decrease the temperature slowly so that the system becomes more and more constrained to exploit low-amplitude movements and, at one point, stable into a low-energy minimum state, which is the solution to the specific problem. Simulated annealing can be applied to many NP-hard combinatorial optimization problems, such as the knapsack problem and the traveling salesperson problem (TSP) [56]. *The Ant Colony Optimization (ACO)* method is inspired by entomology, the science of insect behavior. This follows the same strategy as real ants who lay down pheromones directing each other to resources. The simulated 'ants' (artificial agents) similarly record their positions and the quality of their solutions while exploring their environment so that more ants locate better solutions in later simulation iterations. The ant colony method has been applied to many NP-hard problems, such as routing problems, assignment problems, scheduling problems, telecommunication networks, and industrial problems [57]. Another interesting population-based metaheuristic is Particle Swarm Optimization (PSO), where a swarm of particles (candidate solutions) simultaneously explores a problem's search space to find the global optimum configuration. Because of the simple implementation and few parameter requirements, this approach has been used to solve many applications, such as constrained optimization problems (COPs), min/max problems, multi-objective optimization problems (MOOPs), etc. [58]. Another set of optimization and machine learning techniques are Evolutionary Algorithms (EAs). This work uses a genetic algorithm, one of the best-known and well-proven evolutionary techniques.

### 2.2 Genetic Algorithms

Among the metaheuristic algorithms, the genetic algorithm (GA) is a well-known algorithm that finds its inspiration in the biological processes of evolution. It is based on the mechanics of natural selection and natural genetics. Theoretically, it mimics the Darwinian theory of survival of the fittest [59]. The critical elements of GA are chromosome representation, selection, crossover, mutation, and fitness function computation [60]. The classical approach of a genetic algorithm is illustrated in Figure 1. At first, a population of $n$ chromosomes (e.g., bit strings) is generated randomly, and the fitness of each chromosome is evaluated. The selection step is sometimes called the reproduction operator [61, 62]. Different techniques are used in this step: roulette wheel, rank, tournament, Boltzmann, and stochastic universal sampling, which are well-known in literature. Then, the chromosomes are mixed to create children of the next generation (*offspring*). Many well-known crossover operators exist, such as single-point crossover, two-point, k-point, uniform, partially matched, order, precedence preserving crossover, shuffle, reduced surrogate, cycle, etc. [60]. After that, a uniform mutation operator (e.g., flipping a bit in a child at random [62]) is applied to produce offspring with a small mutation probability to generate mutated offspring so that premature convergence to a local minimum or maximum can be prevented. Finally, the new offspring is placed in the new population. The selection, crossover, mutation, and replacement processes are repeated until a population with a good or accepted fitness value is obtained.

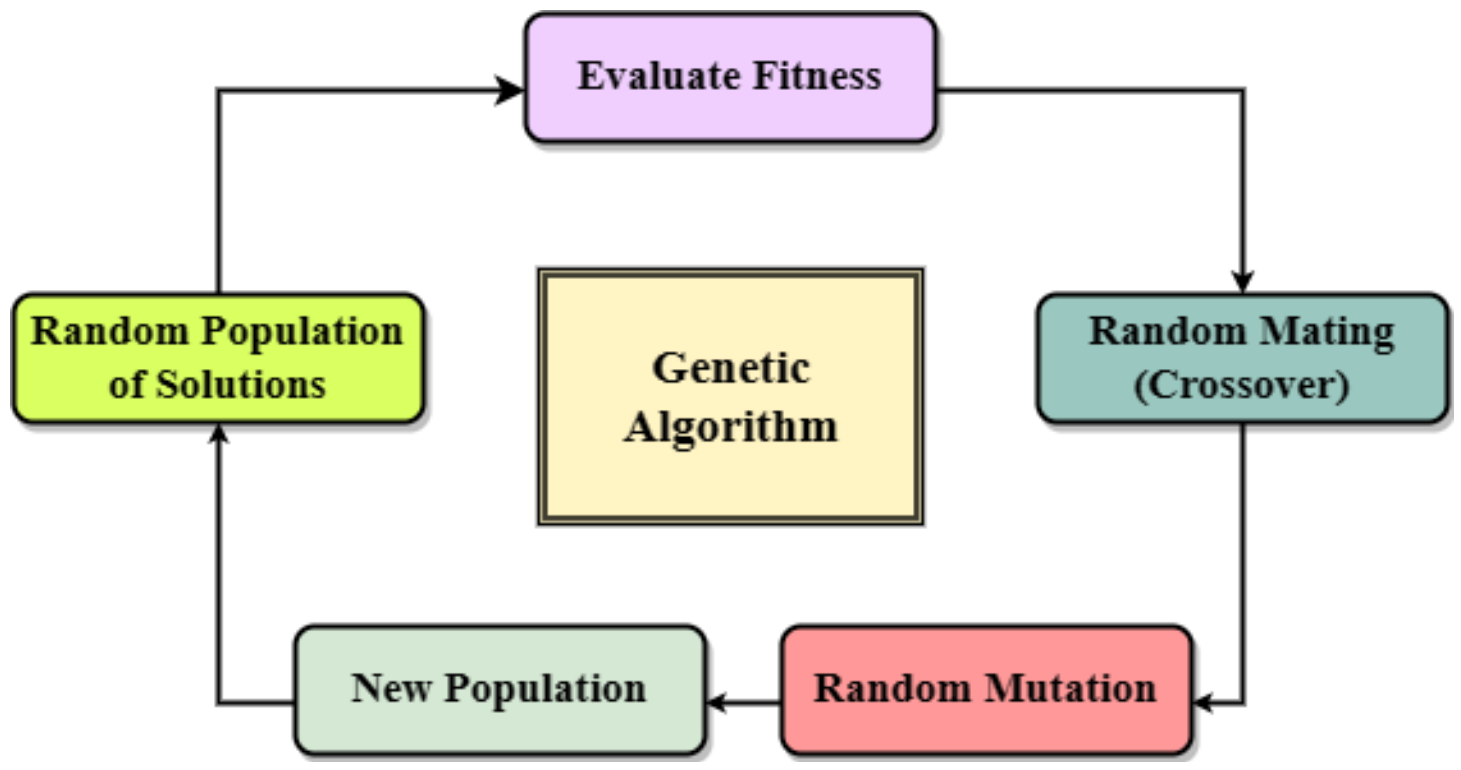

Figure 1. The classical approach of a genetic algorithm.

GAs have been applied as an efficient and high-accuracy metaheuristic for solving various NP-hard problems, ranging from operation management: facility layout problem (FLP) [63], job-shop scheduling (JSS), integrated process planning and scheduling (IPPS) [64] to wireless networking: load balancing [65, 66], bandwidth and channel allocation problem [67, 68], etc. In the information security domain, Kaur and Kumar [69] developed a multi-objective genetic algorithm to optimize the control parameters of the chaotic map, where the secret key was generated using the beta chaotic map. GAs is also used in image processing [70, 71], video processing [72, 73], and medical imaging, such as edge detection in MRI and pulmonary nodules detection in CT scan images [74, 75].

## 3 Proposed Genetic Algorithm

In this work, a genetic algorithm is proposed which is enhanced by three significant features:

- Use census data information to guide the algorithm in making better choices.
- Use of a self-adaptive parameter to adjust the aggressiveness of each reproduction and mutation method
- Use of a parallelized environment to minimize the time spent in the feasibility check control

Using census data improves the choices made by the genetic algorithm: which individuals will reproduce, and which vertices are better choices to minimize the size of a solution (recall that there is a one-to-one correspondence between solutions and individuals). Notably, two distinct census information are proposed. The former (S-Census) is related to individuals. A hash value is computed for individuals to uniquely identify it (if two individuals are associated with the same solution, they are identified as a single individual). Each time a new individual (never identified before) is found, its corresponding hash value is added to the table with value 1, representing that it has been identified only once. If a previously found individual is constructed again, the corresponding value in the hash table is increased. This information is used to reduce the chance of a recurring individual being selected as an ancestral to the next generation, providing greater diversity to the algorithm. More details on the S-Census information are presented in Section 3.3.

The latter census information (V-Census) counts how often each vertex has been included in a solution (individual). The algorithm also uses this information to consider rarely selected vertices, allowing them to be included in dominating sets. Otherwise, the algorithm may preferably include vertices that are immediately good choices, following an unwanted greedy behavior. Both census data are in the decision-making process; however, other information is also considered. To specify the weights of each component in the decision, four parameters are used: $wScensus, wSize, wDegree$ and $wV$ Census. Those parameters are described in detail in Sections 3.3 and 3.4. Also, a parameter is used to specify the aggressiveness of each reproduction and mutation operator, which expresses the expected number of changes each operator will be applied to individuals. This parameter allows the algorithm to escape from locally optimal solutions, avoiding premature termination. The aggressiveness of the algorithm is expressed by the parameter $\delta$, which increases every generation with no improvement in the best solution constructed. Every time an improvement is achieved, the $\delta$ is reset to the starting value. To ensure that the algorithm will not lose reasonable solutions between generations, one-third of the individuals of each generation is copied to the new one. The copied individuals correspond to those with more excellent fitness, and this strategy adds elitist behavior to the algorithm. The remaining two-thirds of the individuals of the new generation are obtained through reproduction from individuals contained in the previous generation. Section 3.4 details this process, including the start value of $\delta$ for each instance. Finally, our algorithm is designed to run in a parallelized environment, reducing the computational cost of checking that individuals corresponds to a feasible solution. As previously explained, the TSS requires multiple iterations to dominate the entire graph, which implies that the algorithm may perform up to $n-1$ iterations to check if a solution is feasible. Since genetic algorithms generally construct many individuals in each generation, the total time required to perform the feasibility check is significant. To reduce this time, the algorithm is implemented using the MPI protocol, dealing with individuals in parallel.

Each processor in the environment is uniquely identified by an integer $p \in \mathbb{N}^*$. At the beginning of the algorithm, the algorithm distributes the task of constructing three new feasible individuals independently to each processor. Next, the generation $B_0$ is constructed from the information provided by the processors. Finally, the algorithm evaluates the individual's fitness in $B_0$, concluding the construction of the initial generation. From a generation $B_i$, the algorithm constructs $B_{i+1}$ as follows. First, each processor selects two parent individuals $P_1$ and $P_2$ from $B_i$. Second, the processor performs the reproduction from $P_1$ and $P_2$, generating two new individuals, $S_1$ and $S_2$. Since the reproduction procedure contains random factors, each processor must ensure independently that $S_1$ and $S_2$ are feasible solutions. Executing this procedure independently on each processor is the goal of the parallelization strategy. Next, each processor p copies the $p$-th best individual $S_3$ from $B_i$. Finally, the new generation is constructed by including individuals $S_1, S_2$ and $S_3$of each processor, and an individual's fitness is evaluated. This process is repeated until the stop conditions are met. Figure 2 illustrates how the parallelization of the genetic algorithm works. In the figure, each processor runs the blue solid boxes in parallel, while the dashed ones represent the task that demands communication and synchronization between processors. It can be noticed that the boxes corresponding to fitness evaluation have both solid and dashed lines since they correspond to two moments: the assessment of the size of each solution in parallel and a broadcast communication to calculate the average and the standard deviation of the solution sizes, required for sigma scaling. The fitness evaluation and sigma scaling [15] are addressed in subsection 3.3.

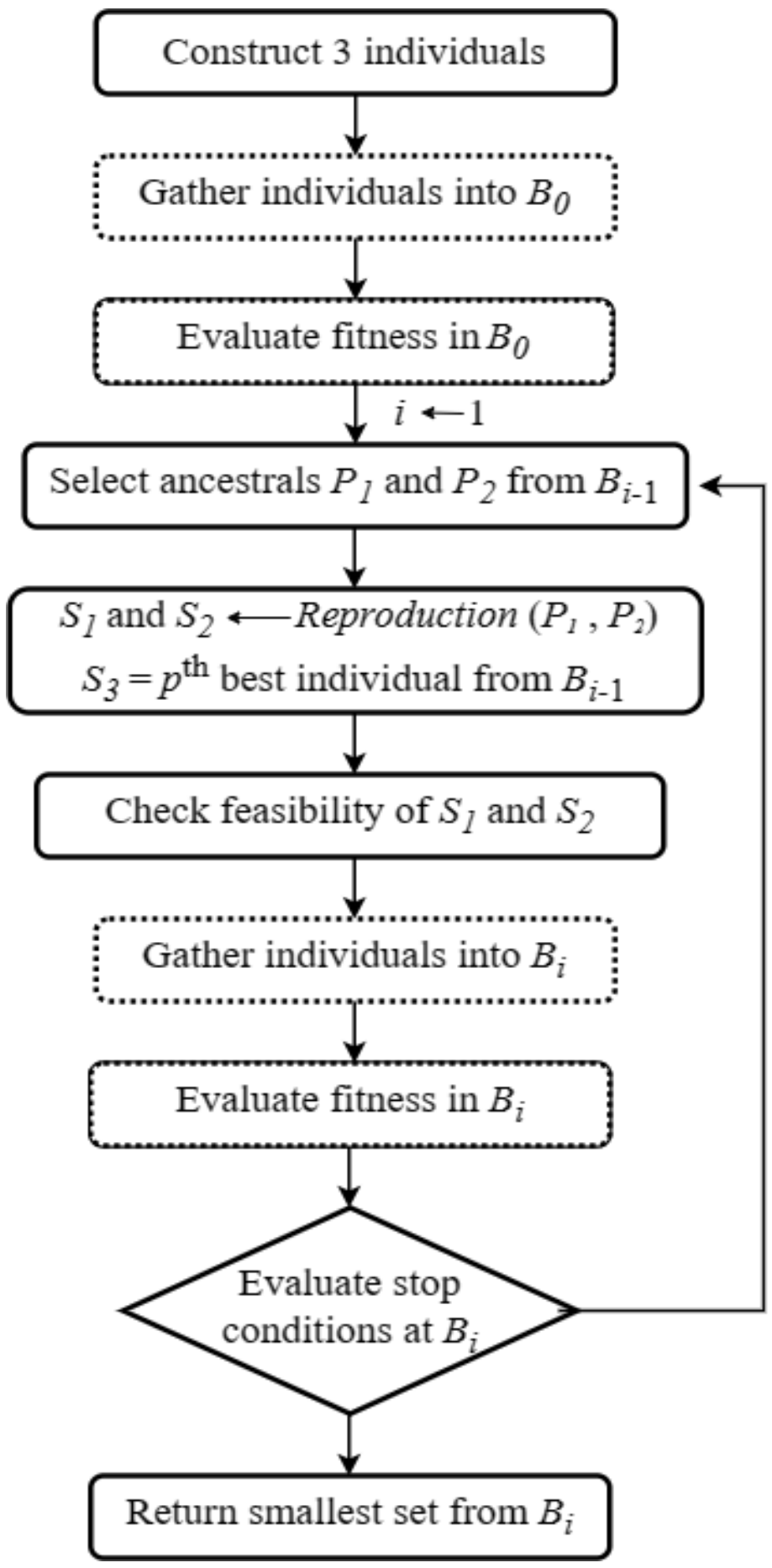

Figure 2. Outline of the parallelization process.

In the following subsections, each algorithm's feature is described in detail. Recall that an individual is a feasible solution (a subset of vertices) that dominates the graph.

### 3.1 Representing an individual

In our approach to the TSS problem, each $S$ is represented as a binary vector $S\,[v]\ v \in V$ such that $S\,[v] = 1$ if and only if $v$ belongs to a set $D \subseteq V$ that is a candidate solution of the TSS. If $D$ is indeed a solution of the TSS, we say that $S$ is a feasible individual or, simply, a feasible solution. Each coordinate $S\,[v]$ is a gene of individual $S$, so genes with value "1" of a feasible individual $S$ map it to a set of vertices $D$ capable of dominating the graph $G$. The use of a binary vector allows to identify if a vertex is included in a solution in $O(1)$ time and accelerates the execution of reproduction methods.

### 3.2 Construction of the initial generation

Each processor $p$ constructs three solutions for TSS. Each solution is constructed using a different method to ensure more diversity in the initial population. The first solution $S_1$ is constructed using a randomized version of the algorithm proposed by Cordasco et al. [6, 7]. Randomness produces a greater diversity of solutions, since the original version is deterministic. Case 3 of "Algorithm TSS" [6] selects the vertex $v$ that maximizes an expression based on the degree and the requirement of each vertex. We changed how vertex $v$ is selected: First, a sorted set is constructed using the same expression; next, a subset of this set is taken with only the $p$ vertices with higher values; finally, $v$ from this subset is randomly chosen. Considering that the first processor has value $p = 1$, this case corresponds to the original version. It is easy to see that the more processors there are, the more diversity is acquired.

A simple randomized greedy algorithm constructs the second solution $S_2$. Algorithm 1 presents this method. The array $S_2$ represents the solution, $N\ [v]$ and $D\ [v]$ store, respectively, the neighborhood and the updated requirement of each vertex $v \in V$. In addition, an auxiliary array $P$ is chosen that keeps track of the propagation of the domination: If $P\ [v] = true$, then $v$ is already dominated. In addition to those arrays, Algorithm 1 also uses a procedure to propagate each addition to the set $S_2$. To propagate the addition of a vertex $x$ to set $S_2$, a queue $Q$ is utilized. First, $Q$ is initialized with $x$. Then, while $Q$ is not empty, at each iteration a vertex $v$ is removed from $Q$, and the neighbors of $v$ are updated, since $v$ is now dominated. If the addition of $v$ dominates a neighbor $u$ of $v$, then $u$ is enqueued in $Q$, which implies that $u$ will be processed later. The function Propagate Domination is described in Algorithm 2.

Finally, the third solution $S_3$ is obtained by mixing the first two: For each vertex $v \in X$, $S_3[v] = S_1[v] \wedge S_3[v]$. Since $S_3$ may not be a feasible individual, it must be checked in order to proceed. First, the domination of each vertex $v \in S_3$ needs to be propagated. During the propagation, unnecessary vertices are also removed from $S_3$, that is, at the moment a vertex $v \in S_3$ is being processed; if its neighbors already dominate it, then it can be removed from $S_3$ without interfering with the domination process. When each vertex of $S_3$ is processed and the propagation finishes, if any vertex is not dominated by $S_3$, new vertices must be added to ensure that $S_3$ is feasible. However, Algorithm 1 already does this since it only finishes when there are no remaining vertices to dominate. Thus, Algorithm 1 is executed from the result of the propagation of $S_3$. When Algorithm 1 finishes, it can be guaranteed that $S_3$ can dominate all vertices. After each processor constructs three individuals $S_1, S_2$ and $S_3$, we can proceed to the next phase: evaluate the quality (fitness) of the constructed solutions.

**Algorithm 1: Randomized Greedy Algorithm for Solution Construction**
**Input:** Graph $G = (V; E)$, Vector of Requirements $R$
**Output:** Solution $S_2$
$X \leftarrow V$
for $v \in X$ **do**
| $S_2\ [v] \leftarrow false\ \ P\ [v] \leftarrow false$
| $D\ [v] \leftarrow R\ [v]$
| $N\ [v] \leftarrow u \in V \mid uv \in E$
**end**
**while** $X \neq \emptyset$; **do**
| Let $BR$ be a biased roulette from $\big(v; w(v)\big)\ \forall\ v \in X$, such that
| $w(v) = \mid N\ [v] \mid / \mid X \mid$
| Select vertex $x$ from $BR$
| $S_2[x] \leftarrow true$
| $PropagateDomination(X; N; D; P; x)$
**end**
**return** $S_2$

**Algorithm 2: Propagation of the Domination of a Vertex**
$PropagateDomination(X; N; D; P; x)$
$Q \leftarrow \{x\}$
**while** $X \neq \emptyset$; **do**
| $v \leftarrow Dequeue\ (Q)$
| **for** $u \in N[v]$ **do**
| | $D\ [u] \leftarrow \max\ (D\ [u] - 1{,}0)$
| | **if** $P\ [u] = false \wedge D\ [u]\ \leq 0$ **then**
| | | $P\ [u] \leftarrow true$
| | | $Enqueue\ (Q, u)$
| | **end**
| | $N\ [u] \leftarrow\ N\ [u] \setminus \{v\}$
| **end**
| $N\ [u] \leftarrow \emptyset$;
| $X \leftarrow X \setminus \{v\}$
**end**
**end**

### 3.3 Fitness Evaluation

To evaluate the quality of each solution, first its size needs to be computed, counting how many 1s are contained in the array. Let $z(S)$ be the size of an individual $S$. During the development of the algorithm, it is noticed that using only the sizes of the individuals as fitness values leads to premature convergence since the best individuals are often selected as ancestors to the next generation. To work around this problem, two improvements in fitness evaluation are implemented: the use of S-Census data and sigma scaling [15]. As previously explained, the idea of S-Census is to keep records of an individual generated by the algorithm and how many times each one has been identified. To keep these records, each time the algorithm finishes a generation, every solution $S$ is analyzed: If $S$ is a solution never identified before, $S$ is added to the database; otherwise, the database is updated to point out that $S$ now has one more occurrence. A hash table is used to store the S-Census database. The key value of an individual $S$ is obtained by converting the binary array into a binary string, where each position is associated with a vertex. It is easy to see that no collisions occur in this hash table. The S-Census database is used to influence the fitness of each solution slightly: Let S-Census $(S)$ be the census of an individual $S$. If the algorithm frequently identifies $S$, the fitness of $S$ is reduced proportionally to S-Census $(S)$. In so doing, if two solutions $S_1$ and $S_2$ have $z(S_1) = z(S_2)$ and if $S_1$ is more frequently identified than $S_2$, the fitness of $S_2$ is greater than $S_1$, increasing the probability of $S_2$ to be selected as an ancestral to the next generation. Two parameters, $wSCensus$ and $wSize$, are utilized to balance the weight of census versus set size during the computation of the fitness of an individual.

Having discussed the factors involved in fitness evaluation, the expression which returns the protofitness of a solution is described. It is important to keep in mind that this expression does not give the final fitness of an individual since the sigma scaling must be applied. Let the protofitness of an individual $S$ be:

$$pf(s) = \frac{\left((n - z(S)) * wSize\right) + ((W - S\text{-}Census(S)) * wSCensus)}{wSize + wScensus} \quad (1)$$

where $W$ is the number of individuals constructed by the algorithm so far. As mentioned before, sigma scaling prevents the algorithm from a premature convergence caused by an accentuated elitism. To do so, sigma scaling formulates the fitness evaluation of an individual as an expression that depends on the quality of the individual as well as the entire generation. Let $B$ and $S \epsilon B$ be, respectively, a generation constructed by the genetic algorithm and a solution contained in this generation. Also, let $\sigma(B)$ and $\overline{pf}(B)$ be the standard deviation and the mean value obtained from the protofitness of an individual in $B$, respectively. The fitness of an individual $S \epsilon B$ is calculated by the following expression:

$$f(S) = \begin{cases} 1, & \sigma(B) = 0 \\ \max\left(1 + \frac{pf(S) - \overline{pf}(B)}{2*\sigma(B)}, 0.01\right) & \sigma(B) \neq 0 \end{cases} \quad (2)$$

Note that $\sigma(B) = 0$ then every individual $S \epsilon B$ has the same protofitness $pf(S)$, and it must be ensured that every individual has the same probability to be chosen as an ancestral to the next generation; in this case, the function always returns the value 1. Since an aleatory factor is also applied in the selection of the ancestry of a generation (presented in Algorithm 6), it is beneficial for solution diversity that every individual has a small probability of being selected. Thus, a small value $a = 0.01$ is defined as the minimum return value for every individual. The evaluation of the protofitness of an individual in a generation can be performed in parallel. However, sigma scaling requires the standard deviation and the average over all protofitness values. Thus, all such values are broadcast to each processor to allow it to calculate the final fitness. This procedure starts as a parallel routine, but only finishes when all processors communicate, demanding a synchronization step.

### 3.4 Construction of a new generation

Once the initial generation is constructed and its fitness is evaluated, we proceed to construct new individuals. Recall that new generations are built until the stop conditions are met. Each processor constructs individuals independently. After such construction, individuals are gathered and broadcasted to all the environment, so that every processor has a local copy of the entire population. This also applies to the initial population. Every processor $p$ (recall that each processor is uniquely identified by its associated label $p$ in the computational environment) constructs three individuals: The first two are obtained by reproduction and mutation; the last one is a copy of the $p$-th best solution found in the previous generation. Considering that the algorithm propagates the best individuals of a generation to the next one, we can state that once a good solution is identified, it is propagated until a better one is found. This elitist behavior ensures that the result is the best solution identified during the execution 1. Another effect of this behavior is that two-thirds of each generation are discarded and replaced by new individuals. The construction of the new two individuals consists of three steps:

- Randomly choose two parents from the previous generation.
- Construct two new individuals using reproduction and mutation.
- Ensure that the new individuals are feasible solutions for the TSS.

Our algorithm uses a biased roulette in which every individual $S$ of the previous generation has a probability $f(S)$ of being randomly chosen. Once the roulette is prepared, two parent individuals $P_1$ and $P_2$ are selected. After defining them, the algorithm proceeds to the reproduction and mutation phase.

#### 3.4.1 Reproduction Operators

To increase the diversity of solutions, fourteen different reproduction methods are implemented

- One point crossover (OPC)
- Two-point crossover (TPC)
- Random crossover (RC)
- Uniform crossover (UC)
- Logical operator AND (AND)
- Logical operator OR (OR)
- Logical operator NOT (NOT)
- Logical operator AND with randomness (R-AND)
- Logical operator OR with randomness (R-OR)
- Average of the previous generation operator (AVG)
- Consensus operator (CO)
- Swap a vertex with its neighbors (SWAP)
- Construct two new individuals (DOUBLE-NEW)
- Forced mutation (FM)

The first two operators are widely found in genetic algorithms literature. Given the two parents $P_1$ and $P_2$, Operator 1 randomly defines an integer $P_2$ in the interval $[2; n-1]$, and the two new children are:

$$S_1 = P_1[1], \dots, P_1[s], P_2[s+1], \dots, P_2[n] \quad (3)$$

And

$$S_2 = P_2[1], \dots, P_2[s], P_1[s+1], \dots, P_1[n] \quad (4)$$

Operator 2 uses two randomly chosen integers $s_1$ and $s_2$ such that $s_1 \epsilon [2, \frac{n}{2} - 1]$ and $s_2 \epsilon [\frac{n}{2} + 1, n - 1]$. The children are defined by the following rule:

$$S_1 = P_1[1], \dots, P_1[s_1], P_2[s_2 + 1], \dots, P_2[s_2], P_1[s_2 + 1], \dots, P_1[n] \quad (5)$$

And

$$S_2 = P_2[1], \dots, P_2[s_1], P_1[s_1 + 1], \dots, P_1[s_2], P_2[s_2 + 1], \dots, P_2[n] \quad (6)$$

Operator 3 follows the same idea used for the previous two but uses a randomized number of cut points. Such number is determined by $\delta$, a parameter of the algorithm. Operator 4 constructs the two children by aleatory swapping positions of the arrays. The expected number of swaps is given by another parameter of the algorithm, $pProbCross$. For each position of arrays $S_1$ and $S_2$, a real number $p \epsilon [0; 1]$ is randomly selected, and the values of each position of $S_1$ and $S_2$ are given by

$$S_1[v] = \begin{cases} P_2[v], & p < pProbCross \\ P_1[v], & otherwise; \end{cases} \quad (7)$$

$$S_2[v] = \begin{cases} P_1[v], & p < pProbCross \\ P_2[v], & otherwise; \end{cases} \quad (8)$$

Operator 5 uses a logical operator AND to define the Boolean values of the arrays associated with children $S_1$ and $S_2$. That is, Operator 5 returns $S_1[v] = S_2[v] = P_1[v] \wedge P_2[v]$. It is important to note that after the construction $S_1$ and $S_2$ are equal, but after possible mutations and after the optimization phase, in which we ensure that both individuals are feasible solutions, random factors involved may change that. Operator 6 follows the same procedure but using the OR operator. Operator 7 constructs two children using the NOT operator, i.e., $S_1[v] = \neg P_1[v]$ and $S_2[v] = \neg P_2[v]$, for each vertex $v$ in solutions $S_1$ and $S_2$. Since Operators 5, 6, and 7 can perform radical changes in the solution, a subtler version (Operator 8) has been developed. In this version, the logical operator is applied only to some vertices of the solution, coping with the other parent values. Operator 8 returns, for each vertex $v$, the values of $S_1[v]$ and $S_2[v]$ as follows:

$$S_1[v] = \begin{cases} P_1[v] \wedge P_2[v], & p < \delta/n \\ P_1[v], & otherwise; \end{cases} \quad (9)$$

$$S_2[v] = \begin{cases} P_1[v] \wedge P_2[v], & p < \delta/n \\ P_2[v], & otherwise; \end{cases} \quad (10)$$

In the above expressions, $p \in [0;1]$ is a randomly selected real number, and $\delta$ is a parameter of the algorithm. Operator 9 follows the same procedure but applies the OR operator. Operator 10 constructs two new individuals based on the average of the solutions contained in the previous generation. That is, if a specific vertex $v$ is contained in most of the individuals from the previous generation, it is assumed that it has a good chance of belonging to a good solution, and thus it is included in a new solution. Two different solutions $S_1$ and $S_2$ are constructed as follows: $v$ is included in $S_1$ if it belongs to at least 50% of the individuals of the previous generation and is included in $S_2$ if this percentage rises to 60%. More formally, if $B_{i-1}$ denotes the previous generation, then:

$$S_1[v] = \begin{cases} true, & if \ \sum_{S_{old} \in B_{i-1}} S_{old}[v] > 0.5 * |B_{i-1}|; \\ false, & otherwise; \end{cases} \quad (11)$$

$$S_2[v] = \begin{cases} true, & if \ \sum_{S_{old} \in B_{i-1}} S_{old}[v] > 0.6 * |B_{i-1}|; \\ false, & otherwise; \end{cases} \quad (12)$$

Operator 11 uses the consensus of all previous generations to determine if a vertex should be included or not in the solution. The consensus operator is applied only at a few random vertices of the solution, so all other positions are copied from the selected parents. To define if there is a consensus about some vertex, the V-Census data is used, which provides some "clues" about which vertices are frequently used. As previously mentioned, the algorithm can use this strategy to make better decisions or even to provide greater diversity by avoiding such vertices. Given a randomly selected real number $r \epsilon [0;1]$ , the value of each position $S_1[v]$ is defined by the following rule:

$$S_1[v] = \begin{cases} true & r < \frac{\delta}{n} \wedge V\text{-}Census[v] > 0.5 * W; \\ false & r < \frac{\delta}{n} \wedge V\text{-}Census[v] \leq 0.5 * W; \\ P_1[v] & otherwise. \end{cases} \quad (13)$$

Recall that $W$ is the total number of distinct individuals already built by the algorithm. The value of $S_2[v]$ follows the same rule, except for using $P_2$ instead of $P_1$. Operator 12 uses characteristics of the TSS to construct new solutions by swapping some vertices. Each new individual is initialized as a copy of its parent ($S_1 \leftarrow P_1$ and $S_2 \leftarrow P_2$) and the operator performs vertex swaps in an individual. As with previous operators, the parameter $\delta$ is also employed to determine how many swaps will be done. Notice that there are two cases in which a vertex v can be dominated: (i) $v \epsilon D_0$ or (ii) $|N[v] \cap D_k| \geq R[v]$, where $D_0$ is the initial set and $D_k$ is the set obtained by propagating the domination rule along $k$ iterations, starting from $D_0$. The idea behind Operator 12 is simple: if both cases (i) and (ii) are true, there is a chance that the removal of $v$ from $D_0$ will not hinder it from being dominated according to case (ii). To avoid the computational cost of obtaining $D_k$ from $D_0$, it is considered that the immediate effects of the domination, that is, the domination of $v$ using only its neighbors included in $D_0$. Thus, all computations are performed on $D_0$.

Algorithm 3 presents the outline of Operator 12. Variables $m_v$ and $c_v$ store, respectively, how many neighbors of $v$ must be added to dominate $v$ (in a single iteration), and the set of neighbors of $v$ that can be included in $S$ to ensure the domination of $v$ (a set of candidate vertices for inclusion). The values of $m_v$ and $c_v$ are initialized, respectively, as $R[v] - |N[v] \cap S|$ and $N[v] \backslash S$. We know that adding $m_v$ neighbors guarantee the domination of $v$ because $N[v] = m_v + |c_v|$ and $R[v] \leq N[v]$.

An essential point about Operator 12 is why the propagation of every change in $S$ is avoided knowing that it may lead to fewer inclusions needed to keep the removed vertex dominated. Note that the propagation function has a high computational cost, and, considering that each processor runs a randomly distributed reproduction method, this extra cost can make all other processors (which may be running faster methods) stop and wait before proceeding together to the next generation.

**Algorithm 3: Operator 12: Swap v for R[v] neighbors of v.**
**Input:** Parents $P_1$ and $P_2$, Parameters $\delta$, Graph $G = (V;E)$, and Vector of Requirements $R$
**Output:** New individuals $S_1$ and $S_2$

**Function** $Operator12(P_1, P_2, \delta, G, R)$
$S_1 \leftarrow Swap(P_1, \delta, G, R)$

$S_2 \leftarrow Swap(P_2, \delta, G, R)$
**return** $S_1$ and $S_2$
**end**

**Function** $Swap(P, \delta, G, R)$
$S \leftarrow P$
**for** $v \in V$ **do**
**if** $S[v] = true$ **then**
Let $r$ be a random float from the interval $[0{,}1]$
**if** $r < \delta/n$ **then**
$S[v] \leftarrow false$
$m_v \leftarrow R[v] \quad c_v \leftarrow \emptyset;$
**for** $u \in N(v)$ **do**
if $S[u] = ture$ then $m_v \leftarrow m_v - 1$ else $c_v \leftarrow c_v \cup \{u\}$ ;
**end**
**for** $i \in \{1 \ldots m_v\}$ **do**
Let $u$ be the vertex with greater degree in $c_v$
$S[u] = ture \;\; c_v \leftarrow c_v \setminus \{u\};$
**end**
**end**
**end**
**end**
**return** $S$
**end**

Therefore, we have opted for a simpler and faster reproduction method. Each operator described so far selects random individuals as ancestors for constructing two new individuals. However, there is one more potential source of good new individuals: Exploit the best solutions of the previous generation. Therefore, two new operators (13 and 14) have been constructed that are based on this idea. Operator 13 still uses randomly selected individuals as parents but considers the size of the best individual of the previous generation to guide the construction of the new individuals. On the other hand, Operator 14 uses the best individuals as the unique source for new solutions. Operator 13 constructs the two new individuals by two distinct methods. The first one constructs an individual that uses parts of each parent to compose a new individual. Given two parents $P_1$ and $P_2$ selected from the previous generation, the new individual $S_1$ is initialized according to the following expression:

$$S_1[v] = \begin{cases} P_1[v], & if\ r = 1 \\ P_2[v], & otherwise \end{cases} \tag{14}$$

where for each $v \in V$ , $r$ is a random integer in the range [1; 2]. This procedure is like the one applied for Operator 4. However, Operator 13 defines a specific size for the constructed solution. Thus, let $P_{best}$ be the individual with greater fitness in the previous generation, and let $t$ be the target size for $S_1$, defined as follows:

$$t = \begin{cases} z(P_{best}) - \delta, & if\ z(P_{best}) > \delta \\ z(P_{best}), & otherwise \end{cases} \quad (15)$$

Recall that $z(P_{best})$ corresponds to the number of vertices in the set represented by $P_{best}$. Given S1 and t, there are three cases: $z(S_1) = 1,\ z(S_1) < t$ and $z(S_1) < t$. In the first case, the algorithm returns $S_1$, since it already has the target size. Otherwise, random vertices must be added or removed, respectively, to match the set size with the target value. After the size correction, $S_1$ is returned, and the algorithm can proceed to the construction of the second individual $S_2$. The construction of $S_2$ follows a different procedure, using the value of V-Census to form a random individual consisting of t vertices, where $t$ corresponds to the same value described in the construction procedure of $S_1$. First, we initialize $S_2[v] = false$ for all $v \in V$ . While $z(S_1) < t$, a biased roulette is constructed formed by pairs $(v, V - Census(V))$ for all $v \in \{x \in V \mid S_2[x] = false\}$ and add a random vertex selected from the roulette into $S_2$. Since every individual constructed by a reproduction method is subjected to the optimization routine, at this moment, the feasibility of the solution is not primely focused, since it will be checked later. As explained in Operator 12, we also chose to avoid the cost of propagating each inclusion in the set. Operator 14 first selects two parent individuals with greater fitness from the previous generation. Let $P_{Best1}$ and $P_{Best2}$ be those individuals. Next, the algorithm executes a forced mutation for each parent to construct new individuals. We call this forced mutation since each solution has only a small chance of suffering a mutation, while Operator 14 always executes it on the two parents.

### 3.4.2 Mutation Procedure

After constructing new solutions, each processor applies random mutations to $S_1$ and $S_2$. The algorithm uses the parameter $pMutation$ to express the probability that an individual is subjected to a mutation. The mutation procedure consists of aleatory changes in the Boolean array to include and exclude vertices from the individual. The value of parameter $\delta$ is used to express how many changes it is expected that the algorithm to perform during the mutation procedure. Algorithm 4 describes this process. Since we are looking for smaller solutions than the previous ones, the probability of removing a vertex from the solution is increased by granting a chance to the swapped value to be false again (line 6) and by applying a second loop only for random vertex removals (lines 8 through 11 of the algorithm).

**Algorithm 4: Mutation Procedure**
**Input:** Individual $S$ and Parameter $\delta$
**Output:** Mutated Individual $S'$
**Procedure** $SMutateUndividual(S, \delta)$

```
    prob_δ ← δ/n
    for i = 1 .... n do
        Let r_1 and r_2 be two random floats in the range [0,1]
        if S[v] = false ∧ r_1 < prob_δ then S[v] ← true;
        if S[v] = ture ∧ r_2 < prob_δ then S[v] ← false;
    end
    for i = 1 .... n do
        Let r be a random float in the range [0,1]
        if S[v] = true ∧ r < prob_δ then S[v] ← false;
    end
    return S_0 = S
end
```

### 3.4.3 Feasibility Check

The next step is to ensure that $S_1$ and $S_2$ are feasible solutions for the TSS. This process is slightly different from the one applied in the construction phase since now we can improve how vertices are chosen to be added or removed from the solutions. Algorithm 5 describes the optimization procedure used to ensure that an individual corresponds to a feasible solution for the TSS. The algorithm behavior is simple: Given the input individual S, a vertex $v$ from $S$ is iteratively chosen and propagated the domination; for each vertex $u$ dominated by this propagation sequence, it is checked if $u \epsilon S$, and if this condition holds then u can be removed from S without changing the capability of u to dominate his neighbors. Once the propagation stops, if there is a non-dominated vertex in the graph then more vertices must be added to correct this. If there are vertices from S that are not dominated yet, we choose one of those vertices and repeat the process. Otherwise, $S$ does not correspond to a dominating set, and new non-dominated vertices must be added from $V \setminus S$ to $S$ to correct this. We keep adding vertices to $S$ until the graph is dominated. Since the algorithm only stops when every vertex is dominated, at the end of the procedure, $S$ is a feasible solution. Recalling that the propagation method is described in Algorithm 2.

**Algorithm 5: Optimization Procedure**
**Input:** Individual $W$, Vertex Census $V\text{-}Census$, Parameters $wDegree$ and $wV\ Census$, Graph $G = (V, E)$ and Vector of Requirements $R$
**Procedure**

```
OptimizeUndividual(S, G, R, V-Census, wDegree, wV Census)
    S' = {v ϵ V |S[v] = true}
    X ← V
    for v ϵ V do
        S[v] ← false    P[v] ← false
        D[v] ← R[v]     N[v] ← {u ϵ X | uv ϵ E}
```

**end**
**while** $X \neq \emptyset$ **do**
$S' \leftarrow S' \setminus \{x \in S' \mid D[x] = ture\}$; // Optimize $S$
**if** $S' \neq \emptyset$ **then**
Construct a biased roulette $BR$ with pairs $(x, w'(x))$, $\forall\, x \in S'$
Randomly select a vertex $v$ from $BR$
$S[v] \leftarrow true$; // Keep $v$ in $S$
**else**
Construct a biased roulette $BR$ with pairs $(x, w'(x))$, $\forall\, x \in X$
Randomly select a vertex $v$ from $BR$
$S[v] \leftarrow true$; // Add new vertex $v$ to $S$
**end**
$PropagateDomination(X, N, D, P, v)$ P
**end**
**end**

Considering that we seek population diversity, an interesting idea is to use the V-Census information to guide the order in which vertices are added during the optimization algorithm, by increasing the probability of vertices that are rarely used to be chosen for the solutions. This idea is applied first to vertices that are already included in the set, but when those vertices are not capable of dominating the graph, another non-dominated vertex is added using this rule. In Algorithm 1, the value of $w(v)$ (used to construct the biased roulette) is formulated as a division of the degree of each vertex by the number of remaining non-dominated vertices. The idea here is to enhance the formulation of $w(v)$ with the V-Census information, thus the probability of a vertex being chosen from the biased roulette is expressed as

$$w'(v) = \frac{\left(\frac{|N[v]|}{|X|} * wDegree + \frac{W - V\text{-}Census(v)}{P} * wV\,Census\right)}{wDegree + wV\,Census} \qquad (16)$$

where $N[v]$ and $X$ are control variables, $W$ is the number of individuals already built by the genetic algorithm, and $wDegree$ and $wV$ Census are parameters of the algorithm. The values of $wDegree$ and $wV\,Census$ define the weights of each value in the probability calculation. An important point is that this procedure seeks for an improvement in solution diversity, which may not lead to immediate improvements in population fitness.

We now have all the elements to fully formulate the reproduction process, described in Algorithm 6. Once each processor p constructs three feasible solutions (two by reproduction and one by copying a previous good solution), the algorithm gathers the new individuals into a new generation $B_{new}$ and broadcast $B_{new}$ to every processor. After that, the fitness of an individual in $B_{new}$ is measured. Since each processor executes the method independently, each reproduction operator is implemented to spend a similar amount of time.

**Algorithm 6: Reproduction Process**
**Input:** Previous generation $B$, Parameters $\delta$ and $pMutation$
**Output:** Three individuals $S_1$, $S_2$ and $S_3$
**Function** $Reproduction(B, \delta, pMutation)$
Let $r$ be a random integer selected from the interval [1,14]
**if** $r \leq 13$ **then**
Let $\phi_r$ be the Reproduction Operator $r$
Let $BR$ be a biased roulette with pairs $(S, f(S)), \forall S \epsilon B$
Randomly select two individuals $P_1$ and $P_2$ from $BR$
**if** $r < 13$ **then**
$(S_1, S_2) \leftarrow \phi_r (P_1, P_2)$
**else**
Let $P_{best}$ be the individual with greater $f(S) \forall S \epsilon B$
$t \leftarrow P_{best} - \delta$
**if** $t \leq 0$ **then** $t \leftarrow z( P_{best})$;
$(S_1, S_2) \leftarrow \phi_{13} (P_{Best1}, P_{Best2}, t)$
**end**
**else**
Let $P_{Best1}$ and $P_{Best2}$ be two individuals with greater $f(S), \forall S \epsilon B$
Let $\phi_{14}$ be Reproduction Operator 14
$(S_1, S_2) \leftarrow \phi_{14} (P_{Best1}, P_{Best2})$
**end**
Let $r_1$ and $r_1$ be two random real numbers in the range [0,1]
**if** $r_1 < pMutation$ **then** $MutateIndividual(S_1, \delta)$;
**if** $r_2 < pMutation$ **then** $MutateIndividual(S_2, \delta)$;
$OptimizeIndividual\ (S_1)$; $OptimizeIndividual\ (S_2)$
Let $S_3$ be the $p$-th best individual from $B$
**return** $S_1, S_2$ and $S_3$
**end**

### 3.4.4 Stop Conditions

The algorithm constructs new generations until the stop conditions are met. When the algorithm stops, it returns the smallest solution in the last generation. Three parameters are leveraged to express the stop conditions: $gMin, gMax$ and $gWImprovement$. The first two represent the minimum and the maximum number of generations, respectively. The third one specifies how many generations the algorithm constructs with no improvement in the best solution found until it stops.

### 3.4.5 Outline of the proposed genetic algorithm

Having discussed all the necessary elements of our algorithm, we describe in Algorithm 7 its complete formulation.

**Algorithm 7: Outline of the Proposed Genetic Algorithm**
**Input:** Graph $G$, vector $R$, parameters
$gMin, gMax, gWImprovement, pMutation, wSize, wSCensus, wDegree, wV\ Census$
**Output:** Solution $G$ for $G$ and $R$
$i \leftarrow 0$; $ctImprovement \leftarrow 0$; Calculate $\delta_0$ from $G$ and $R$
$\delta_{step} \leftarrow (3 * \delta_0)/\ gW\ Improvement$
Construct starting generation $B_0$
Evaluate the fitness of every individual $S \epsilon B_0$
Update populational census: S-Census and V-Census
**do**
  $i \leftarrow i + 1$
  From $B_{i-1}$, construct a new generation $B_i$ by reproduction
  Evaluate the fitness of every individual $S \epsilon B_i$
  Update populational census: S-Census and V-Census
  **if** $\min(z(S)\ \forall\ S \epsilon B_i) < \min(z(S')\ \forall\ S' \epsilon B_{i-1})$ **then**
    $ctImprovement \leftarrow 0$; $\delta \leftarrow \delta_0$
  **else**
    $ctImprovement \leftarrow ctImprovement + 1$
    $\delta \leftarrow \delta_0 + ctImprovement * \delta_{step}$
  **end**
**while** $(i \leq gMin \vee ctImprovement \leq gWImprovement) \wedge i \leq gMax$;
**return** $S \mid z(S) \leq z(S')\ \forall\ S, S' \epsilon B_i$

## 4 Results

In this section, the self-adaptive aggressive nature of the proposed algorithm is evaluated with a detail of the possible values of each parameter. Besides, the computational environment used in our tests is presented. Furthermore, the results of our experiments are exhibited when compared to the results of Cordasco et al. [6, 7]. We have followed the same format used in [6, 7] to present the results, splitting the results into two sub-sections. The first one presents the results obtained in random graphs with 30 and 50 vertices, while the second exhibits the results obtained for fourteen graphs that map large real-life social networks.

### 4.1 Self-adaptive aggressiveness of the algorithm

As previously explained, the proposed genetic algorithm can adapt itself by increasing or reducing the extension of changes performed in the reproduction and mutation phases. The aggressiveness of the algorithm is expressed by the parameter $\delta$, which is proportionally adjusted to how many generations the algorithm constructs with no improvement in solution size. The initial value of $\delta$ is calculated from the input graph and the vector of requirements using the following expression:

$$\delta_0(G,R) = \min\left(\frac{(\max(R[v] \forall v \in V))^2 * n}{\sum_{v \in V} R[v]}, \frac{n}{4}\right) \quad (17)$$

The value of $\delta$ expresses a reasonable number of changes in an individual $S$ that can be performed without transforming $S$ into a completely different individual. For small graphs, the value of $S$ is usually bounded by $n = 4$. Table 1 shows some examples of values of $\delta_0$ calculated for some instances presented in Section 3. In each instance, the requirement of each vertex $v \in V$ is defined by $\min(\deg(v), R)$, where $R$ is the value of the column "Requirement".

Table 1. Examples of values of $\delta_0$ for some instances

| **Graph** | **Vertices** | **Edges** | **Requirement** | $\delta_0$ |
|---|---|---|---|---|
| ca-GrQc | 5,241 | 14,484 | 1 | 1 |
| ca-GrQc | 5,241 | 14,484 | 5 | 8 |
| ca-GrQc | 5,241 | 14,484 | 10 | 24 |
| ca-HepTh | 9,875 | 25,973 | 1 | 1 |
| ca-HepTh | 9,875 | 25,973 | 5 | 7 |
| ca-HepTh | 9,875 | 25,973 | 10 | 23 |
| BlogCatalog | 88,784 | 2,093,195 | 1 | 1 |
| BlogCatalog | 88,784 | 2,093,195 | 5 | 7 |
| BlogCatalog | 88,784 | 2,093,195 | 10 | 17 |
| Douban | 154,907 | 327,162 | 1 | 1 |
| Douban | 154,907 | 327,162 | 5 | 14 |
| Douban | 154,907 | 327,162 | 10 | 46 |
| YouTube2 | 1,138,499 | 2,990,443 | 2 | 2 |
| YouTube2 | 1,138,499 | 2,990,443 | 5 | 11 |
| YouTube2 | 1,138,499 | 2,990,443 | 7 | 20 |

Additionally, Figure 3 shows an average time consumption chart for each operator, including: (a) the time needed for the reproduction, and (b) the additional time for the optimization procedure. Since each processor executes the method independently, each reproduction operator is implemented to spend a similar amount of time, which reduces the overall waiting time. Such charts have been obtained during the execution of the experiments presented in the following subsections.

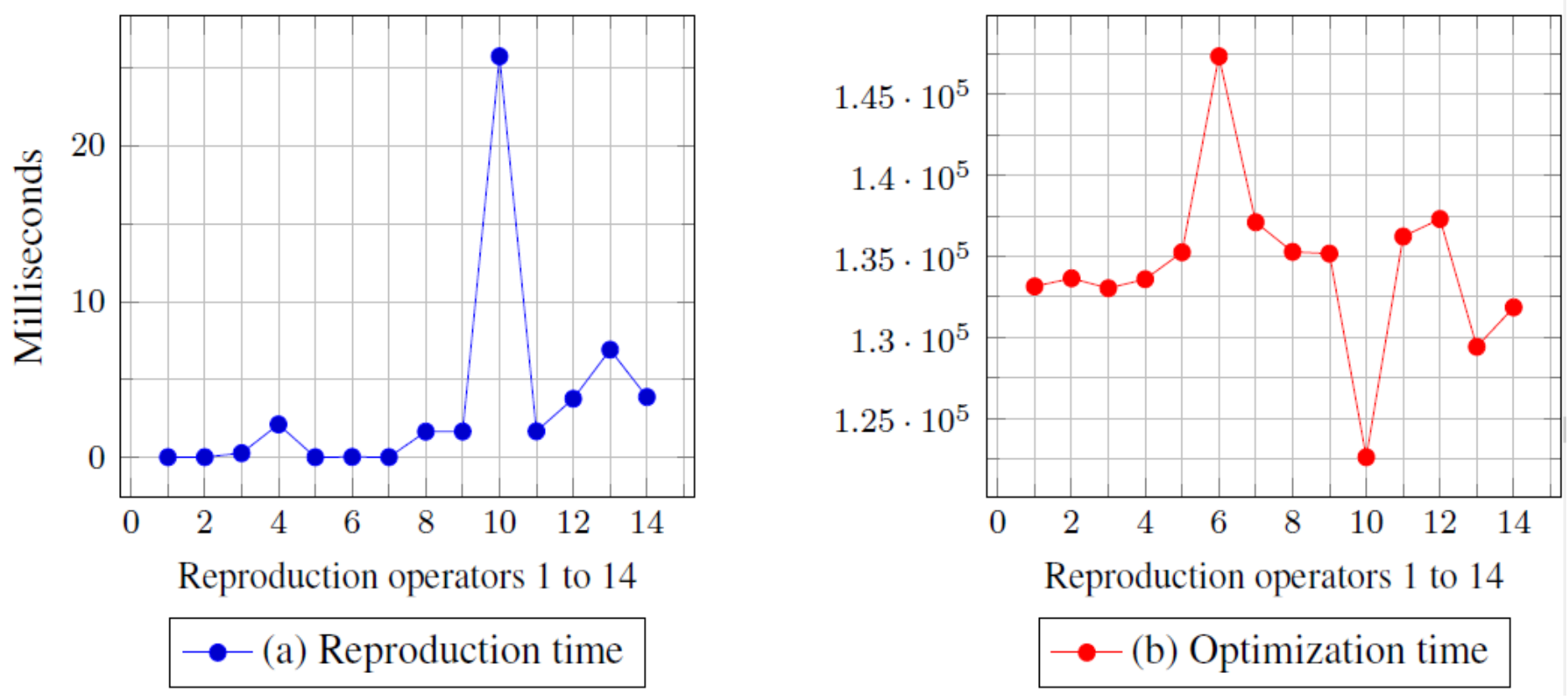

Figure 3. Time consumption for each reproduction operator.

### 4.2 Algorithm Parameters and Computation Environment

The value associated with each parameter is defined by a series of comparisons in several instances. The values described here correspond to the best balance between result and time consumption. As previously explained, the value of parameter $\delta$ is obtained from the instance data and is not given as a fixed input parameter. First, we discuss the parameters used in the fitness evaluation: $wSCensus$ and $wSize$. As explained, these parameters specify the weights associated with the S-Census data and the size of the solution. We defined $wSize = 0.98$ and $wSCensus = 0.02$. Next, parameters $wDegree$ and $wV\ Census$ are used to define the weights associated with the V-Census data, combined with the degree of each vertex to calculate the probability of a vertex to be selected and added to the solution, during the optimization procedure. It has been empirically observed that setting $wDegree = 0.98$ and $wV\ Census = 0.02$ results in a good equilibrium of weights. During the reproduction phase, it is also needed to calibrate parameters $pProbCross$ and $pMutation$. Parameter $pProbCross$ is used in Operator 4 to define the probability of swapping the values associated with a vertex in two iterations, while $pMutation$ specifies the probability of mutation of an individual. It is empirically defined that the best values for those parameters are $pProbCross$=0.3 and $pMutation = 0.025$. The last parameters of the algorithm are related to the stop condition: $gMin, gMax, gWImprovement$. The first two define, respectively, the minimum and the maximum number of generations executed by the genetic algorithm, while parameter $gWImprovement$ defines how many generations the algorithm will execute with no improvement in the best-solution size. Considering that those parameters have a significant impact in execution time, it is defined that $gMin = 10$, $gMax = 500$, and $gWImprovement = 50$. An important point about the stop parameters is that the algorithm will perform one more iteration than the specified since the first generation $B_0$ is excluded from the counting of the generations. Thus, $gMin = gMax = 1$ will lead to two generations $B_0$ and $B_1$. The same applies to $gWImprovement$, as can be noticed in Subsection 4.3. We can now proceed to the computational environment. Our tests are executed using the Compute Engine Service, provided by Google Cloud Platform [16]. We used one dedicated computer-optimized (C2) instance with 60 processors and 240 GB of RAM. Furthermore, the instance runs Ubuntu Linux 2018.04 LTS, and the Open MPI protocol [17] is used to enable the parallelization of our algorithm, which was implemented in C++.

### 4.3 Results from Random Graphs

The same procedure applied by Cordasco et al. [6, 7] is followed to test our algorithm. The initial tests consist of instances formed by random graphs and random requirements. The graphs were generated using a parameter $q$, which expresses the probability of the existence of an edge. Thus, to construct a graph $G(n, q)$ with $n$ vertices with probability $q$, we iterate through every possible edge $e$ of the graph and generate a random real value $r$. If $r < q$, then $e$ is added to the graph. After constructing the graph, its connectivity is evaluated and disconnected graphs are discarded since they represent two distinct instances with fewer vertices. It is easy to see that greater values of q produce denser graphs. After constructing the graph, the requirement vector is defined. For each vertex $v \in V, R[v]$ corresponds to a random integer in the interval $[1, \deg(v)]$, where $\deg(v)$is the degree of each vertex. The genetic algorithm was then applied to graphs with 30 and 50 vertices, with probabilities $q \in 0.1, 0.2, \ldots\ldots, 0.9$. Considering that, for smaller instances, the optimal solution can be obtained, our results are presented by comparing with the optimal result for each instance. We first tried to formulate the TSS as an Integer Linear Programming problem, as proposed in [6, 7] and execute it using IBM ILOG CPLEX Optimization Studio [18]. However, it is noticed that the time to process each instance was significantly high. Thus, an exact method is implemented for the TSS using a parallel backtracking algorithm. Three characteristics are developed to improve this backtracking algorithm:

- Solutions corresponding to ordered sequence of vertices are only considered, that is, a sequence $(v_i, v_{i+1}, \ldots., v_k)$ is analyzed, while $(v_{i-1}, v_i, \ldots., v_k)$ is ignored (assuming that $v_i < v_{i-1}, \ \forall \ v_i, v_{i+1} \in V$);
- At each iteration, non-dominated vertices are only considered as candidates for being included in the solution, since there is no reason to add an already dominated vertex.

- Finally, since a parallelized environment is used for the genetic algorithm, the same environment is utilized to speed up the backtracking algorithm. To decrease the need for communication between processors, a starting point is sent to each processor, which corresponds to the first vertex that must be included in the solutions analyzed by this processor. Since only ordered sequences of vertices are considered, if a processor receives, e.g., the vertex with label $v_3$ as its starting point, it will discard solutions that contain vertices with labels $v_1$ and $v_2$.

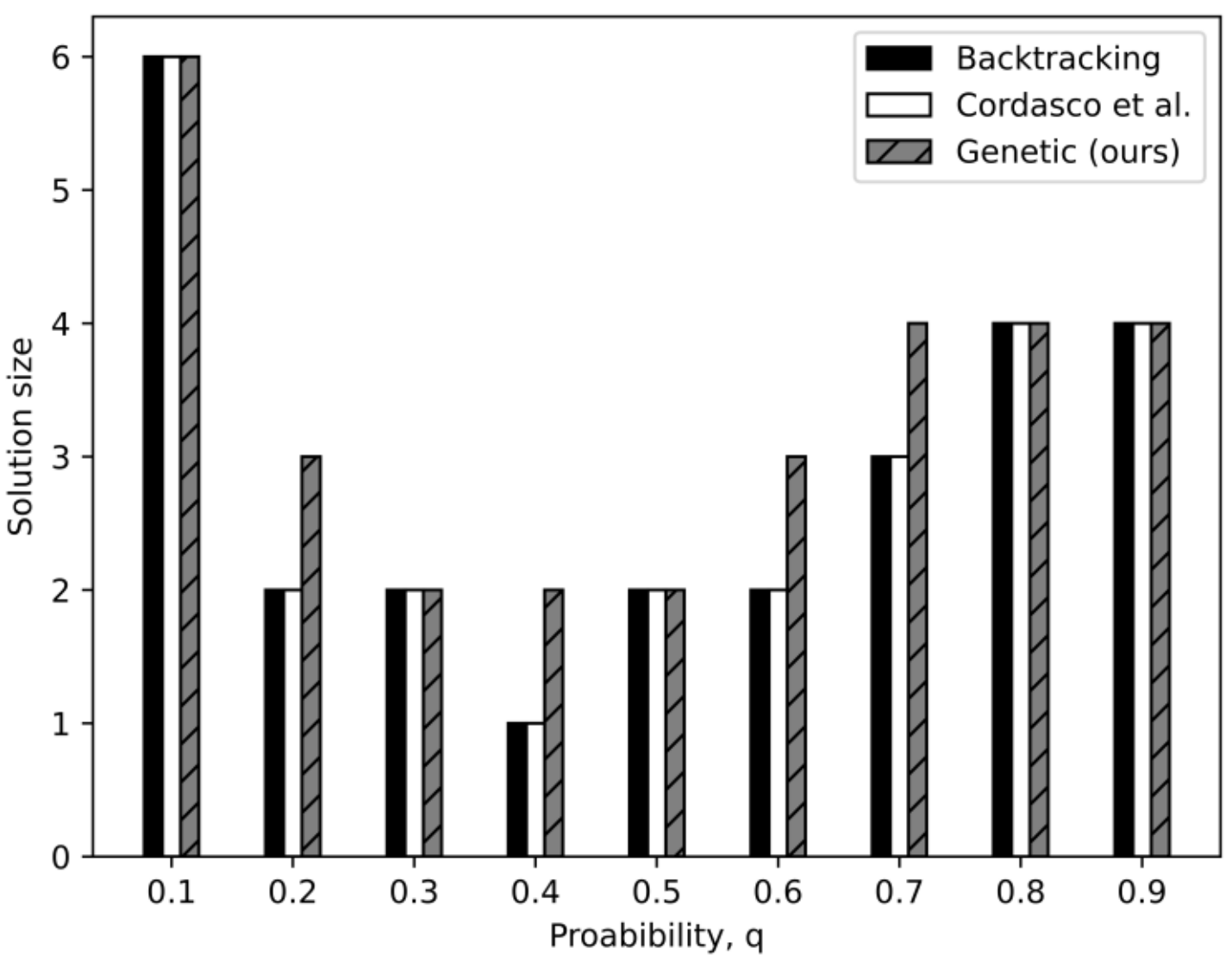

Figure 4. Experiments with random graphs with 30 vertices.

Figures 4 and 5 present the results obtained for random graphs with 30 and 50 vertices, respectively. In both figures, the black column represents the results obtained by the backtracking algorithm, the gray column shows the results obtained by the proposed genetic algorithm, and the black column the results from an implementation of the algorithm by Cordasco et al. [6, 7]. For each probability, one trial is executed. In all cases, our algorithm finds the optimal solution, while in five cases (q=0.2, 0.3, 0.4, 0.6, 0.7), the previous best algorithm ([6, 7]) misses the optimal solution.

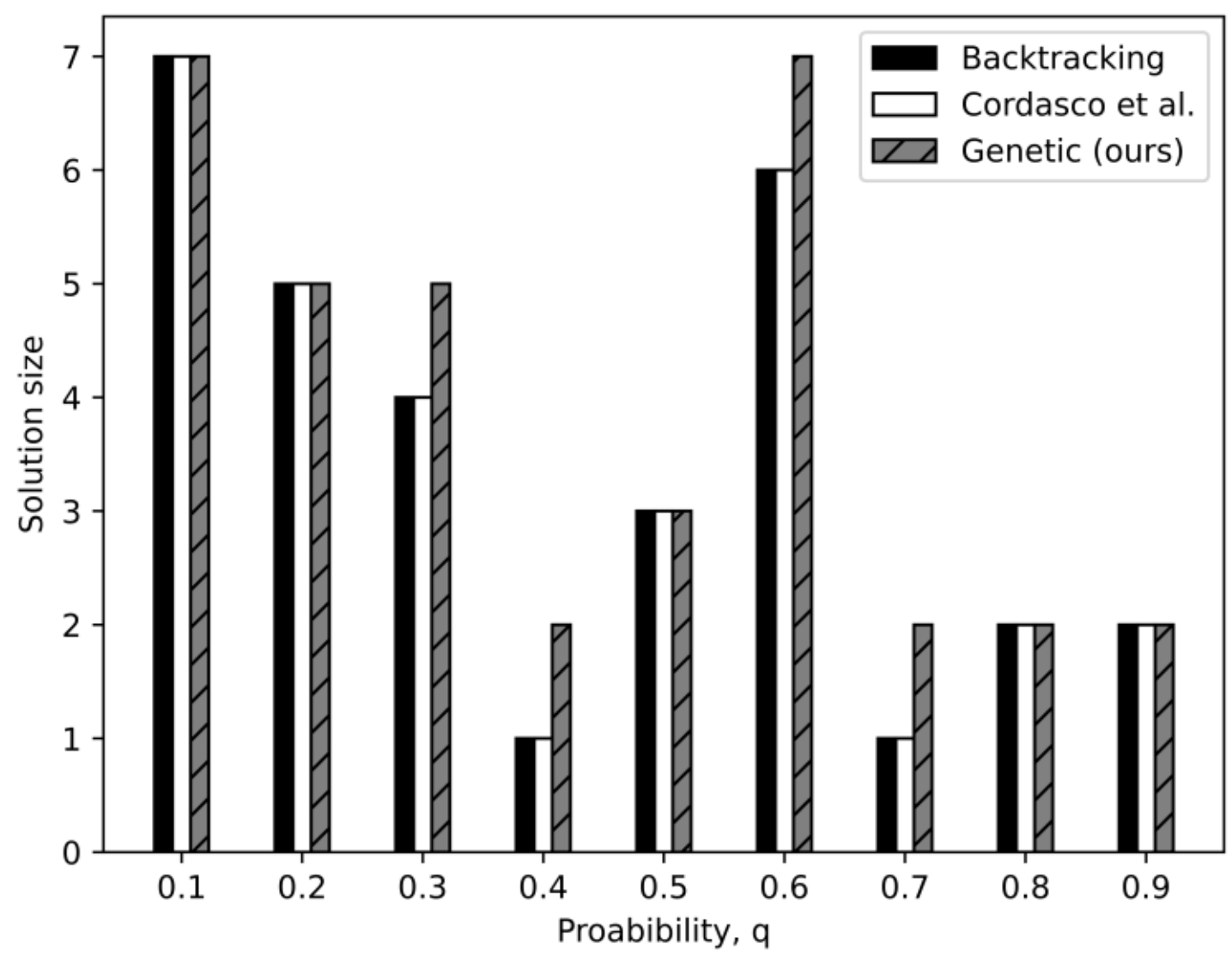

Figure 5. Experiments with random graphs with 50 vertices.

### 4.4 Results from Large Real-life Social Networks

Following the experiment setting proposed by Cordasco et al. [6, 7], we have also executed our algorithm on graphs associated with large real-life networks. Fourteen different graphs are explored, with sizes from 5,241 to 1,138,499 vertices, obtained from two sources. The first one is the Stanford Large Network Dataset Collection (SNAP) [13] and consists of five graphs: ca-AstroPh, ca-CondMat, ca-GrQc, ca-HepPh, and ca-HepTh. The second source is the Social Computing Data Repository from Arizona State University [14], with nine graphs: BlogCatalog, BlogCatalog2, BlogCatalog3, BuzzNet, Delicious, Douban, Last.fm, Livemocha, and Youtube2. The first step to execute our algorithm on those graphs was to process them into binary files and then feed them into our algorithm. However, in this step, there are some inconsistencies in the graphs. For example, the instance Delicious is provided as two files that contain the vertices and the edges of the graph. The file with the edge contains only 1,385,843 entries, but the description says that the graph has 1,419,519 edges. Thus, there are missing edges, as stated in the dataset website. Therefore, to detect and correct inconsistencies, all graphs are preprocessed before starting the tests with the genetic algorithm. This pre-processing step consists of three actions: removal of $multi\ edges$ (two or more edges with the same end vertices), removal of $loop$ (edges starting and ending at the same vertex), and removal of $isolated\ vertices$ (vertices with degree 0), resulting in a simple undirected graph. Table 2 presents the processing results.

Table 2. Pre-processing of instances.

| Instance | Instance data | | Processing Result | | |
|---|---|---|---|---|---|
| | **Vertices** | **Edges** | **Vertices** | **Edges** | **Δ** |
| BlogCatalog | 88,784 | 4,186,390 | 88,784 | 2,093,195 | 9,444 |
| BlogCatalog2 | 97,884 | 2,043,701 | 97,884 | 1,668,647 | 27,849 |
| BlogCatalog3 | 10,312 | 333,983 | 10,312 | 333,983 | 3,992 |
| BuzzNet | 101,168 | 4,284,534 | 101,163 | 2,763,066 | 64,289 |
| ca-AstroPh | 18,772 | 396,160 | 18,771 | 198,050 | 504 |
| ca-CondMat | 23,133 | 186,936 | 23,133 | 93,439 | 279 |
| ca-GrQc | 5,242 | 28,980 | 5,241 | 14,484 | 81 |
| ca-HepPh | 12,008 | 237,010 | 12,006 | 118,489 | 491 |
| Delicious | 103,144 | 1,419,519 | 102,154 | 881,594 | 2,473 |
| Douban | 154,907 | 654,188 | 154,907 | 327,162 | 287 |
| Last.fm | 108,493 | 5,115,300 | 108,493 | 3,470,597 | 5,225 |
| Livemocha | 104,438 | 2,196,188 | 104,103 | 2,193,083 | 2,980 |
| Youtube2 | 1,138,499 | 2,990,443 | 1,138,499 | 2,990,443 | 28,754 |

After pre-processing the instances, the data is stored into binary files. Those binary files were given as input for the genetic algorithm, together with the stop conditions. We encountered a problem with the YouTube2 instance, whose size is approximately fifteen times larger than the average of all other instance sizes. In our initial attempts to run the genetic algorithm on YouTube2, in some cases the algorithm did not stop after more than 24 hours of processing. Therefore, a particular stop condition is established for this case, consisting of a timer that stops the processing after 20 hours of execution (discarding the reading time of the binary file). The time limit of 20 hours is defined by observing the amount of time spent in each previous instance, divided by instance size. This value was then multiplied by the size of the YouTube2 instance. Table 3 shows the average value of each instance. Columns "Average C.size" and "Average C.time" present the results obtained from the work by Cordasco et al. [6, 7], executed on the same computational environment. Columns "Average G.size" and "Average G.time" contain the results obtained by our proposed genetic algorithm. Columns "Average C.time" and "Average G.time" present the execution time in minutes, disregarding instance reading time. Finally, columns "Average Number of Gen" and "Average Improv." contain, respectively, how many generations the genetic algorithm performed for each instance and the difference between "Average C.size" and "Average G.size".

Regarding results on large real-life networks, we have improved the previously best-known solution in *almost* all cases (except BlogCatalog2), with an average improvement of 9.57 vertices and a total improvement of 134 vertices. As the algorithm is implemented by Cordasco et al. [6, 7] and executed in the same computational environment, an accurate time comparison is also established. Disregarding the YouTube2 instance, which is an exceptional case, their algorithm runs the remaining thirteen instances in 1.1 minutes on average, while ours spends 3.51 minutes. Though it can be acknowledged that the algorithm proposed by Cordasco et al. [6, 7] provides better time efficiency, our higher time consumption is worth the investment, since the solutions returned by genetic algorithm are never overcome and are better in most cases. Besides, this difference is expected since the proposed genetic algorithm produces multiple generations that require a significant amount of computational power.

Table 3: Comparative analysis of the Genetic Algorithm and the algorithm by Cordasco et al. [6, 7], for each instance.

| **Instance** | **Average C.time (min)** | **Average G.time (min)** | **Average Number of Gen.** | **Average C.size** | **Average G.size** | **Average Improv.** |
|---|---|---|---|---|---|---|
| BlogCatalog | 1.0783 | 5.3567 | 51.1000 | 599.9 | 599.8 | 0.1000 |
| BlogCatalog2 | 3.0000 | 4.3850 | 51.0000 | 12.4 | 12.4 | 0.0000 |
| BlogCatalog3 | 0.0217 | 3.9800 | 56.2000 | 27.8 | 27 | 0.8000 |
| BuzzNet | 1.6133 | 4.2733 | 62.8000 | 357 | 354.3 | 2.7000 |
| ca-AstroPh | 0.0367 | 1.7067 | 72.6000 | 2528.7 | 2524.3 | 4.4000 |
| ca-CondMat | 0.0717 | 2.4683 | 70.7000 | 5701.4 | 5693.7 | 7.7000 |
| ca-GrQc | 0.0000 | 3.2517 | 64.7000 | 1682.3 | 1677 | 5.3000 |
| ca-HepPh | 0.0117 | 3.5550 | 71.2000 | 2348.8 | 2343 | 5.8000 |
| ca-HepTh | 0.0017 | 3.2117 | 74.8000 | 2720.1 | 2714.8 | 5.3000 |
| Delicious | 0.7133 | 1.0417 | 71.9000 | 1422 | 1399.4 | 22.6000 |
| Douban | 3.5517 | 5.7667 | 60.8000 | 1736.8 | 1732.8 | 4.0000 |
| Last.fm | 2.7250 | 4.4600 | 66.8000 | 104.4 | 102.2 | 2.2000 |
| Livemocha | 1.3000 | 2.2000 | 59.4000 | 1024.1 | 1022.6 | 1.5000 |
| YouTube2 | 114.6733 | 150.0000 | 5.6000 | 155063.8 | 154992.2 | 71.6000 |
| **Total** | 128.7984 | 195.6568 | 839.6 | 175329.5 | 175195.5 | 134.00 |
| **Overall Average** | 1.08 (except YouTube2) | 3.51 (except YouTube2) | 59.9714 | 12523.535 | 12513.964 | **9.5714** |

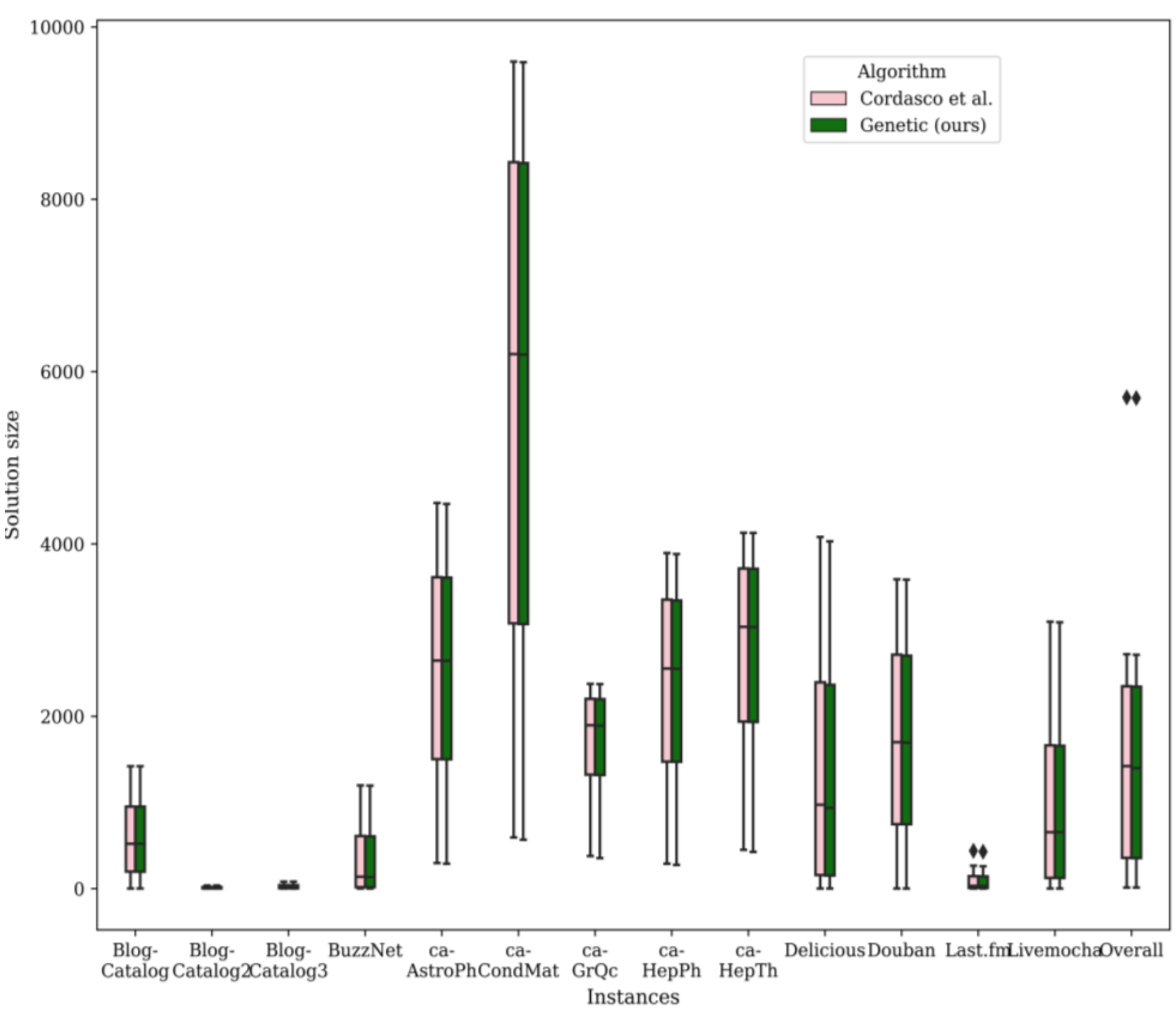

Figure 6. Comparison of our proposed genetic algorithm and algorithm by Cordasco et al. [6, 7] on solution size.

The comparison of our proposed genetic algorithm and algorithm by Cordasco et al. [6, 7] is depicted in Figure 6, on the solution size metric. Though we have got around 71 vertices improvement in the YouTube2 instance, this is not considered in the box plot so that the effect can be better visualized. It is known that the TSS problem is a minimization problem, where the aim is to find a minimum set of vertices, satisfying some required conditions. In our experiment, better results are obtained (than [6, 7]) by reducing the solution (set) size, and Figure 6 proves our claim. Recently, Raghavan and Zhang [76] proposed a branch-and-cut strategy for solving weighted target set selection problems. Though this approach performs well, it is computationally inexpensive while experimenting with large graphs. Besides, since they devised an exact algorithm, the approach is not flexible for dynamic networks. On the contrary, our proposed metaheuristic-based technique is well-suited for any environment since it gives near-optimal solutions in a considerate time.

In Figure 7, the impact of each operator on the collection of the best new solutions found is presented. It is worth noticing that each best new solution can result from several improvements, each one achieved by the application of an operator. In Figure 6, it is observed that Operator 5 (AND) has a significantly higher impact, roughly 22 times, than Operator 6 (OR). This can be explained by observing that the AND operator leads to smaller solutions than the OR operator, since the former adds a vertex to the resulting solution only if both parents agree about such vertex (both must have true assignments), while in the latter a vertex is added if only one parent contains such vertex in the corresponding set. In addition, operators DOUBLE-NEW, CO, AVG, R-AND, and R-OR largely impact the reproduction of solutions, having 9.7, 10.8, 8.7, 9.9, and 9.0 percentile, respectively. Furthermore, other operators occupy average portions in the pie chart, indicating a decent impact on producing the best new solutions.

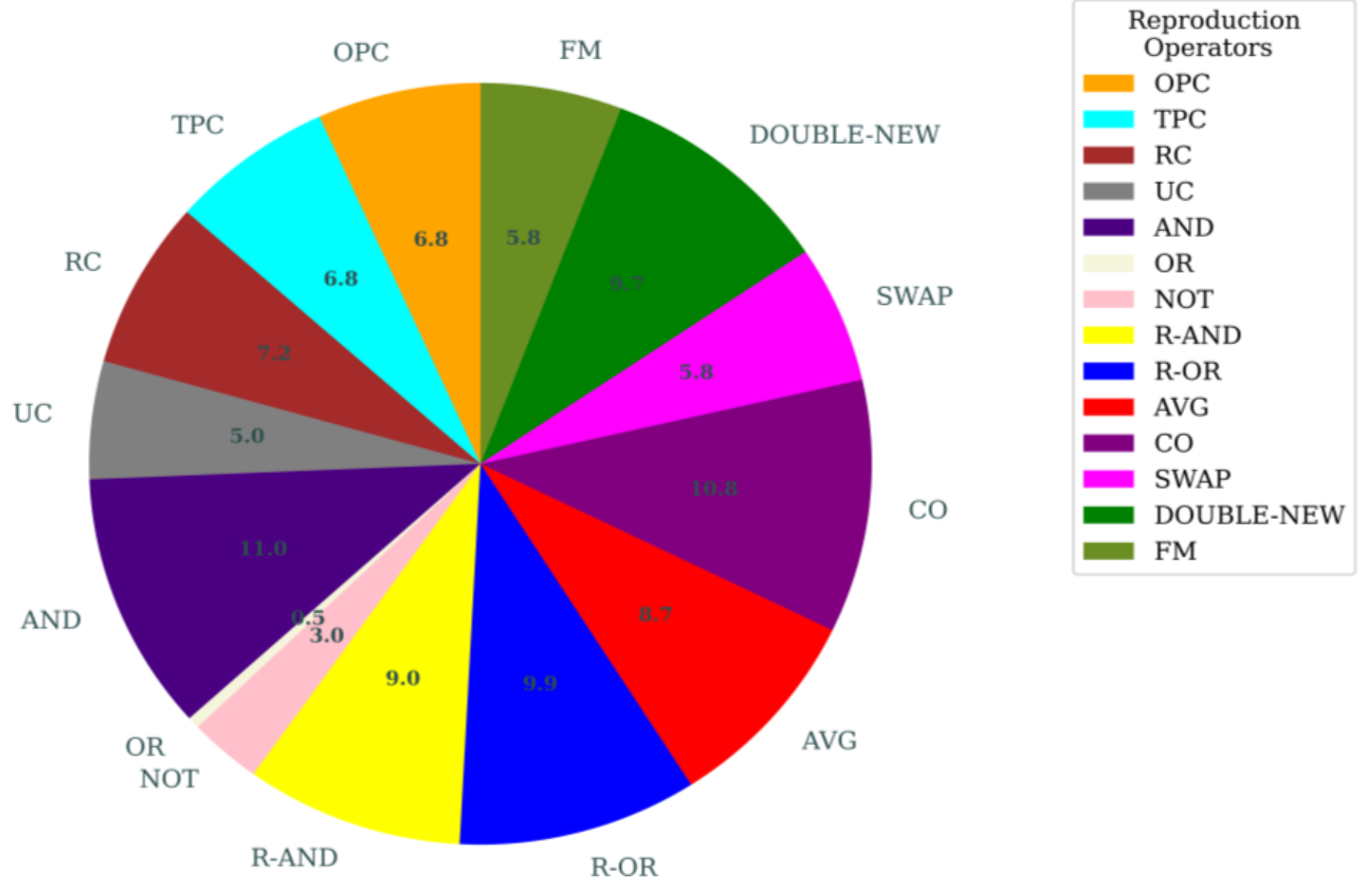

Figure 7. Impact of operators on new solutions.

### 4.5 Comparison with Raw Genetic Algorithm and other Discussions

***Comparison***. Raw Genetic Algorithms (RGA) are prone to premature convergence, especially in complex problems like the Target Set Selection (TSS), where the search space is vast, and the fitness landscape may have many local optima. The sooner the population converges to a suboptimal solution, the more challenging it is to find a better solution. RGA struggles to maintain diversity in the population over time, especially if the mutation rate is low. This lack of diversity can lead to stagnation, where the algorithm stops improving even though better solutions may exist. In contrast, the census operator in the Census-Based Genetic Algorithm (CBGA) helps maintain diversity by aggregating information across the entire population. This global perspective allows the algorithm to avoid getting stuck in local optima and continue exploring promising regions of the search space. CBGA is more likely to converge to high-quality solutions because the census operator guides the population toward better solutions without losing diversity. This makes CBGA more robust in complex optimization problems like TSS.

In RGA, the convergence speed is often slow due to the reliance on random mutations and crossovers. The algorithm may take many generations to explore the search space effectively, especially in problems with large solution spaces like TSS. The convergence speed of RGA is highly sensitive to parameter settings. Poorly chosen parameters like low mutation rate and small population size can lead to slow progress or premature convergence. On the other hand, the census operator in CBGA uses global information to accelerate convergence by guiding the search. CBGA uses aggregated data from the population to make informed decisions about the solutions to explore next rather than depending on random mutations and crossovers. CBGA's ability to maintain diversity and highlight promising regions of the search space leads to faster convergence without getting stuck in local optima.

***Applications***. The census-based genetic algorithm is a versatile optimization framework that can be adapted to various problems in graph theory and social networks. The key feature of CBGA, the census operator, aggregates global information from the population to guide the evolutionary process, making it particularly well-suited for problems involving complex graphs or networks in which maintaining diversity and avoiding premature convergence are critical. Some of these potential applications are discussed below.

*Community Detection.* To partition a network into communities (clusters) such that nodes within a community are densely connected, while connections between communities are sparse. The census operator can aggregate information about which nodes are frequently grouped across the population, helping identify stable community structures. This allows the algorithm to explore possible partitions, avoid local optima, and find more accurate community structures.

*Graph Coloring.* For the graph coloring problem, a minimum number of colors are constrained such that no two nearby nodes can have the same color. The census operator determines which specific nodes are assigned to certain colors across the population and balances the coloring process. This allows the CBGA to explore different coloring strategies, which helps to avoid local optima and reduce the number of colors used.

*Network Resilience and Robustness.* To identify critical nodes or edges whose removal would significantly disrupt the network's connectivity or functionality. The census operator can track how often specific nodes or edges are targeted for removal across the population, helping identify the most critical components.

*Link Prediction.* To predict missing or future links in a network based on existing topology and node attributes. The census operator can aggregate information about which node pairs are frequently predicted to form links across the population, guiding the search toward likely candidates. It ensures that less obvious but potentially valid links are also considered by maintaining diversity.

*Resource Allocation in Networks.* Allocating limited resources, such as budget, vaccines, or bandwidth, across nodes or edges in a network maximizes utility or minimizes cost. The census operator measures the frequency of the allocated resources for each specific node or edge in the population, ensuring optimal configurations. It also ensures that alternative resource allocation strategies are explored, avoiding suboptimal distributions.

Although the Census-Based Genetic Algorithm (CBGA) can be a powerful tool for solving optimization problems in social networks and graph theory, still a few issues need to be solved in future. One of them is the computational overhead of census operators and fitness evaluation, which can become prohibitive for extremely large networks. The proposed CBGA algorithm is susceptible to its parameter settings, which may sometimes require high tuning for different types of networks. Despite these limitations, CBGA remains a promising approach for many network-related problems, provided that appropriate modifications and optimizations are made to address the specific characteristics of the network and problem at hand.

## 5 Conclusion

In this paper, a novel genetic algorithm is proposed, which uses census information produced during the execution to improve the algorithm's decision policy. The goal was to avoid premature convergence and enhance solution diversity to escape from local optima. The experimental results, of 14 large graphs that map real-life social networks, have proved that our algorithm comes up with around 9.57 vertices improvements (on average) and 134 vertices improvements (in total) over the previously best-known solutions, indicating the goal was successfully accomplished and establishing the use of census information as a good additive for genetic algorithms. Importantly, our results have outweighed the previous best-known solutions for random graphs and more suitable for dynamic and complex network graphs. Moreover, we have also discovered a few new reproduction operators, for instance swapping a vertex with its neighbors.

Since data required to implement the use of census information is always available in genetic algorithms, collecting and using such data is not a difficult task. However, memory used to store census information can be a drawback, especially in problems which large solution representations. In future, extended research can be accomplished to make our algorithm more memory and time efficient.

**Conflict of Interest:** The authors declare that they have no conflict of interest.

**Data Availability:** The dataset will be made available upon reasonable request.

**Funding:** This research received no external funding.